\documentclass{article}

\usepackage{arxiv}

\usepackage[utf8]{inputenc} 
\usepackage[T1]{fontenc}    
\usepackage{hyperref}       
\usepackage{url}            
\usepackage{booktabs}       
\usepackage{amsfonts}       
\usepackage{nicefrac}       
\usepackage{microtype}      
\usepackage{lipsum}
\usepackage{graphicx}
\usepackage{subfigmat}
\usepackage{amsmath,amssymb}

\newcommand{\bm}[1]{{\mbox{\boldmath $#1$}}}

\title{Computation harvesting in road traffic dynamics}

\author{
  Hiroyasu Ando, T. Okamoto, H. Chang, T. Noguchi\\
  Division of policy and planning sciences,\\
  University of Tsukuba\\
  1-1-1 Tennoudai, Tsukuba, 305-8573 JAPAN \\
  \texttt{ando@sk.tsukuba.ac.jp} \\
   \And
  Shinji Nakaoka\\
  Department of Biology\\
  Hokkaido University\\
  Sapporo, Hokkaido, 060-0808 JAPAN \\
  \texttt{snakaoka@sci.hokudai.ac.jp} \\
}

\begin{document}
\maketitle

\begin{abstract}
Owing to recent advances in artificial intelligence and internet of things (IoT) technologies, collected big data facilitates high computational performance, while its computational resources and energy cost are large. 
Moreover, data are often collected but not used.  
To solve these problems, we propose a framework for a computational model that follows a natural computational system, such as the human brain, and does not rely heavily on electronic computers. 
In particular, we propose a methodology based on the concept of ‘computation harvesting’, which uses IoT data collected from rich sensors and leaves most of the computational processes to real-world phenomena as collected data. 
This aspect assumes that large-scale computations can be fast and resilient. 
Herein, we perform prediction tasks using real-world road traffic data to show the feasibility of computation harvesting. 
First, we show that the substantial computation in traffic flow is resilient against sensor failure and real-time traffic changes due to several combinations of harvesting from spatiotemporal dynamics to synthesize specific patterns. 
Next, we show the practicality of this method as a real-time prediction because of its low computational cost. Finally, we show that, compared to conventional methods, our method requires lower resources while providing a comparable performance.

\end{abstract}


\section{Introduction}
Recent developments of internet of things (IoT)/artificial intelligence (AI)/big data technologies have enabled an increase in communication capacity and speed, creating global challenges of saving computational resources and reducing power consumption. 
In particular, deep learning, which is the fundamental technology for AI, processes large amounts of data and computational resources \cite{Garcia2019}. 
Although it may be possible to save computational time for deep learning by improving the algorithms, such as few-shot learning \cite{Koch2015, Vinyals2016, Wang2020}, its computational process still depends on the performance of digital computers. Accordingly, 5G communication will further strain computational resources along with its power consumption.
On the other hand, the energy consumption of the brain in terms of electricity is equivalent to approximately 550 Wh per day for an adult male \cite{Mergenthaler2013}. This amount of power consumption is only enough to run a single GPU for a few hours; however, this is not sufficient for information processing by the brain in daily life. 
The energy efficiency of the brain's information processing is due to utilisation of physical and chemical phenomena as its computational resources, which is very different from that of electronic computers. 
From a resource perspective, processing various information in the real world using digital computers alone requires a huge amount of computational 
resources. 
Therefore, it is worthwhile to introduce an alternative computational resource.

It is known that there are various computational resources in nature, including the brain's information processing. For example, there is a framework called natural computation \cite{Dodig-Crnkovic2013}, in which natural phenomena are regarded as computational resources for information processing in applications such as quantum computation \cite{Deutsch1985, Kane1998}, cellular computation \cite{Daniel2013, Singh2014}, DNA computation \cite{Adleman1994}, optical computation \cite{Ibrahim2004}, and amoeba computers \cite{Aono2007}. 
The natural computation proposes and implements a computational model for utilising natural phenomena as a part of the computational resources. Furthermore, the idea is that natural phenomena themselves are regarded as information processing, i.e., that they are occurring as a result of some type of computation \cite{Zenil2013}. 
Based on the perspective that real-world phenomena perform some computation, we use those natural phenomena for a desired computation. Specifically, we propose a half-natural and half-artificial computational method that takes advantage of the computational power of natural phenomena used as they are, which is applicable to various types of real-world phenomena.

Meanwhile, reservoir computing is similar to the above-introduced concept, which was originally proposed in terms of artificial neural networks. It has been shown that prediction and classification tasks utilising various physical phenomena can be performed by using reservoir computing \cite{Jaeger2004, Luko2012}. 
The principle of reservoir computing takes advantage of the dynamics of nonlinear systems, where the response to an input to a reservoir, i.e. nonlinear dynamical systems, is mapped in a high-dimensional space. Thereafter, it allows complex tasks to be accomplished by a linear model of the response. In particular, this is called a physical reservoir and has recently been intensively studied \cite{GT2019}. 
The physical reservoir is regarded as computation by natural phenomena accompanied by artificially designed experimental systems; i.e., the implementation of natural phenomena into the reservoir is externally controllable \cite{Brunner2013, Nakajima2018}. 
In other words, similar to electronic computers, the boundary conditions of the physical reservoir need to be set externally.
In this study, our interest is to consider a simple computation based on the principle of reservoir computing, while avoiding setting boundary conditions for natural phenomena as much as possible.
This follows the idea that the first thing to do for natural phenomena is to observe them, and the next thing to do is to consider the observed time series as the response of a nonlinear system. Thereafter, a complex task is performed in a linear model of the response.

From the viewpoint of observing natural phenomena and extracting a linear model from the observation, we call this framework `computation harvesting' (CH).
Although the physical reservoir can also be viewed as a part of computation harvesting, it requires additional artificial costs to construct experimental systems. 
The computation harvesting explained herein is less expensive by considering the cost of the artificial part. 
Furthermore, it is important to note that a pattern of harvesting is not necessarily unique because methods of observing natural phenomena vary, and a single purpose of computation can be achieved with various ways of harvesting; for example, some sensor failures can be compensated by other sensors. This implies that CH should be resilient, and it is also considered as a type of parallel processing. 
In this study, to exemplify the concept of CH, we focus on the traffic dynamics on a road network.
The evaluation task for the proposed idea is predicting traffic volumes, in which observed traffic dynamics is also used for the prediction.
The principle of the artificial part of CH is based on reservoir computing. 
We show the feasibility of the proposed concept with real traffic data.

The organisation of this paper is as follows. Section 2 provides a schematic explanation of the concept of CH. 
Section 3 presents the results of validation on real data. In particular, traffic predictions outside the system are realised from the traffic dynamics within the observed system.
In Section 4, we discuss the real-world implementation of the proposed method and provide a final summary and future research directions. In the Appendix, the interpretation of the reservoir computing model of the artificial part is described for its implementation in the road traffic network system.

\section{Basic concept of computation harvesting}

\begin{figure}
\begin{minipage}{\hsize}
 \centering
  \includegraphics[width=\textwidth]{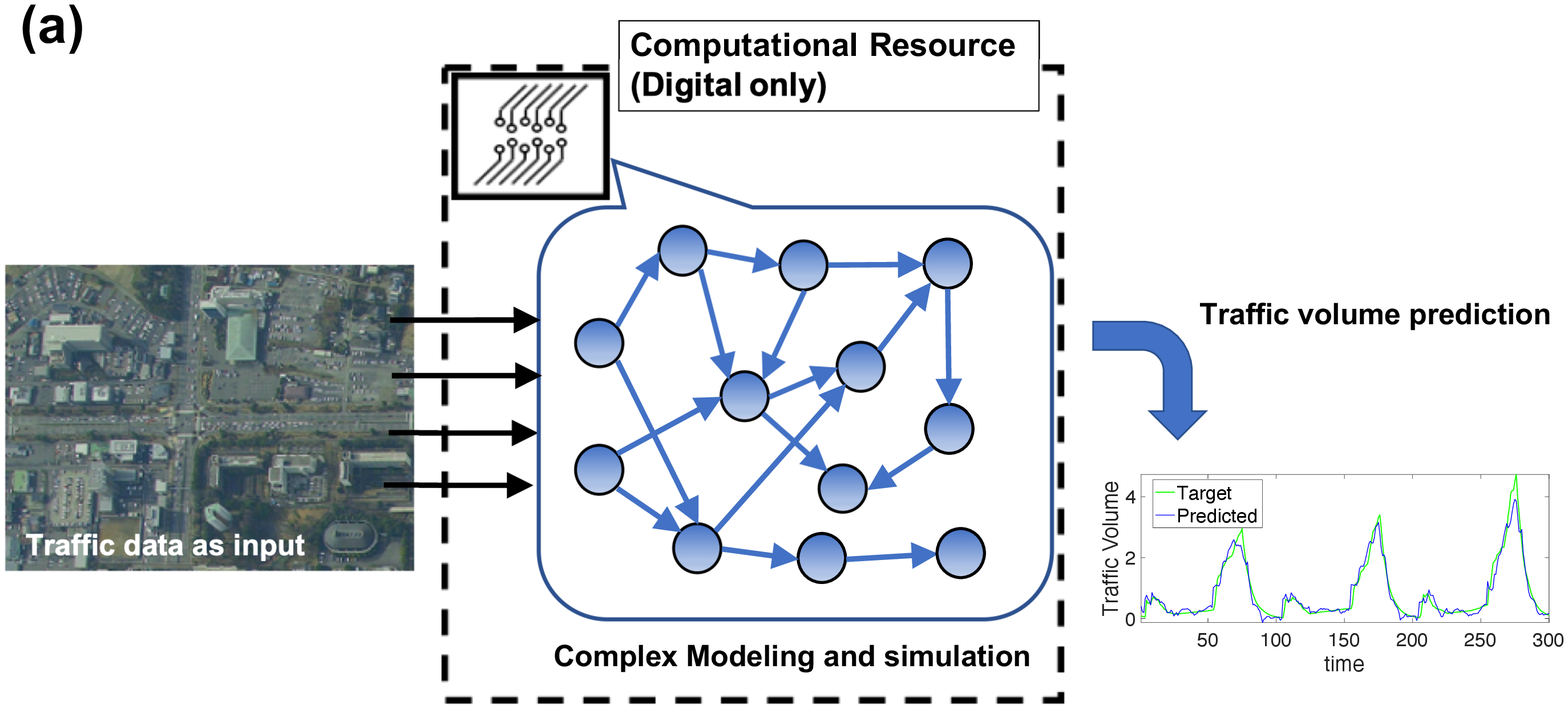}
 \end{minipage}
 \begin{minipage}{\hsize}
 \centering
  \includegraphics[width=\textwidth]{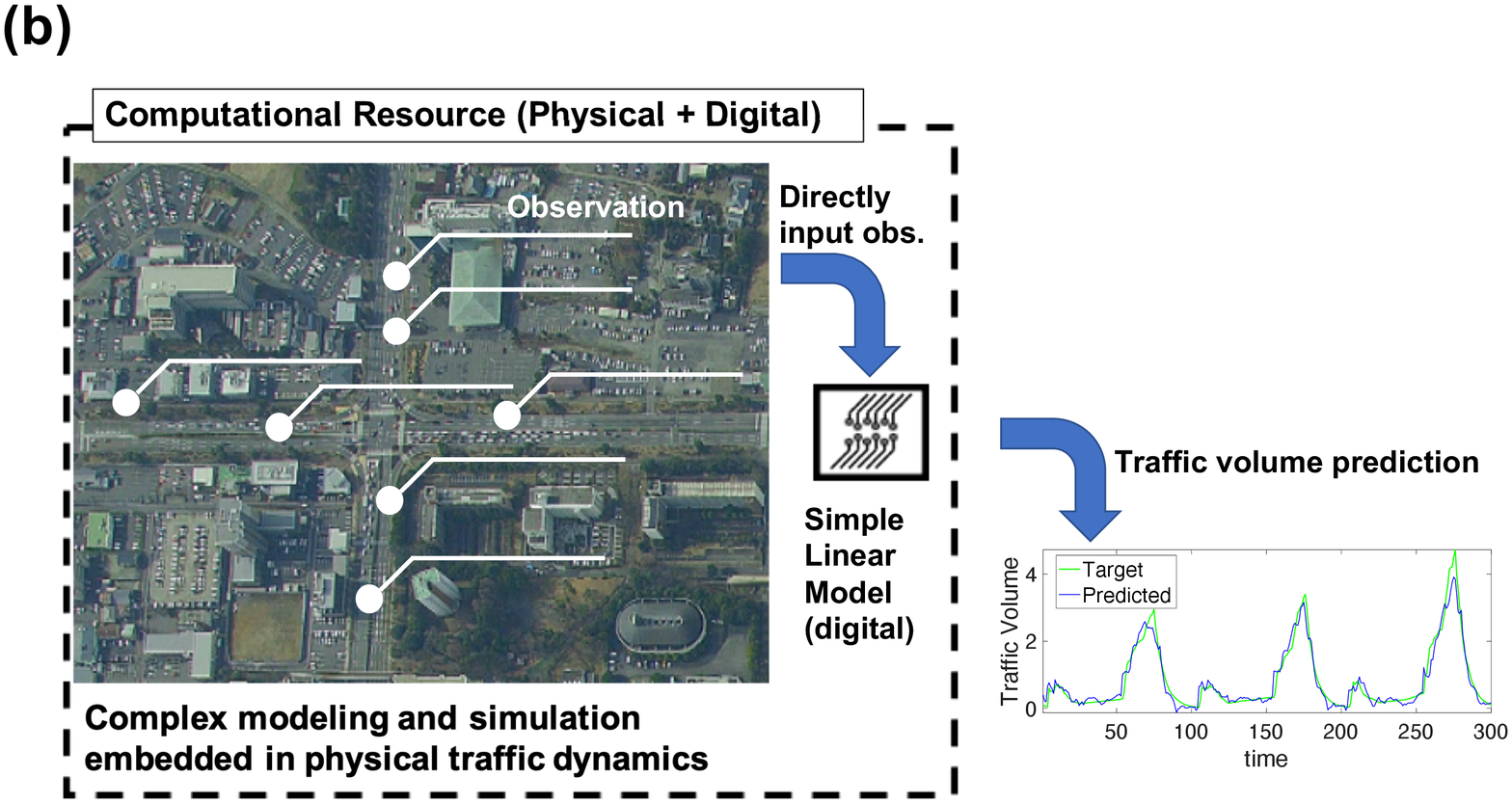}
   \end{minipage}
  \caption{Schematic illustration of CH from road traffic dynamics. (a) Conventional scheme for predicting traffic volume by digital computers alone. Data of the traffic are inputted to the computers. (b) Proposed scheme for prediction in which the traffic dynamics itself is involved in the computational resources. The contribution from digital computers can be small.}
  \label{fig:fig1}
\end{figure}

Here, we describe the concept of CH with respect to predicting traffic volume from the dynamics of vehicle movement in the transportation network. 
Figure \ref{fig:fig1} (a) shows the framework of the traffic prediction task with conventional digital computers alone. As the figure indicates, data obtained from a real traffic flow are inputted to digital computers and are processed with some learning models for the prediction of traffic volume. 
In contrast , Fig. \ref{fig:fig1} (b) shows the proposed framework; real-time traffic data, i.e. dynamics of traffic flow, are observed as a part of modelling and simulation for the prediction task. 
Subsequently, information extracted from the real traffic dynamics is inputted to digital computers. 
The computational resources depend on real-world phenomena and are shared with digital computers. 
Therefore, the cost of the digital computer is lower than that of the conventional framework. 
Note that the computational costs by the digital part in the conventional prediction, such as deep learning, are intensive, and that in the proposed one, such as linear regression, are less intensive. By interpreting the CH as reservoir computing, the liner regression corresponds to the learning output weights.

\subsection{Prediction model of computation harvesting in traffic dynamics}

To exemplify the concept of CH from traffic flow, 
we consider a task of traffic volume prediction in terms of 
a linear combination of traffic volume time series observed at multiple locations.  
In general, to predict the traffic dynamics, microscopic models of each vehicle and traffic signals 
at intersection are required, and they are simulated by computers with many vehicles in the transportation network. 
However, we are only interested in the observed time series of motion of vehicles in the network and 
do not consider the microscopic model of vehicles in real world. 
In other words, the real-world phenomena are exploited as a result of the naturally occurring simulation by real traffic dynamics. 
Additionally, to complete the prediction task, we introduce a simple linear model of target time series in which the time series observed at multiple locations correspond to variables. This linear coupling is based on the idea of reservoir computing \cite{Jaeger2004, GT2019}. 
The time series observed at multiple locations corresponds to the dynamics of reservoir units in the framework of reservoir computing.

To apply the idea to real-world traffic, the prediction task of traffic volume can be formulated as follows. 
We assume that the transportation network is modelled by a lattice with nodes as the traffic signals and links as the road between the traffic signals. Basically, the links are directed, and there are two directions between two nodes. 
Here, we denote a link or road by index $i$, and the traffic volume of link $i$ at time $t$ is denoted by $x_i(t)$. The number of links is $N$. 

When the traffic volume of a target link $\hat{i}$ is $x_{\hat{i}}$, we predict the time series:
\begin{align}
\bm{y}=[x_{\hat{i}}(t_0+\tau),x_{\hat{i}}(t_0+\tau+1),\ldots, x_{\hat{i}}(t_0+\tau+T-1)], 
\end{align} 
where $\bm{y}\in \mathbb{R}^{1\times T}$. $\tau$ is the forecasting horizon and $T$ is the length of the observed time series. 
$\bm{y}$ is predicted from a set of time series $\bm{X}$ by observing other links.  
\begin{align}
\bm{X}=[X_1,X_2, \ldots, X_{\hat{i}-1},X_{\hat{i}+1},\ldots, X_{N}], 
\end{align}
where $X_i=[x_i(t_0),x_i(t_0+1),\ldots ,x_i(t_0+T-1) ]^\top$.   
The learning phase for the prediction output from $\bm{X}$ is conducted by a linear model, i.e. $\bm{y}=W_{out}\bm{X}$. 
The output weight $W_{out}$ is estimated by the ridge regression as 
\begin{align}
W_{out}=\bm{y}\bm{X}(\bm{X}\bm{X}^\top+\beta \bm{I})^{-1}. \label{traffic_comp}
\end{align}
$\beta$ is a hyper parameter, and we consider a pseudo-inverse matrix in the estimation of $W_{out}$. 
In the next section, we apply this model to real-world traffic flow and predict traffic volume in order to verify the principle of CH. Note that the prediction of traffic volume is the task of CH.

\section{Experimental verification}

\begin{figure}
  \centering
  \includegraphics[width=8.0cm]{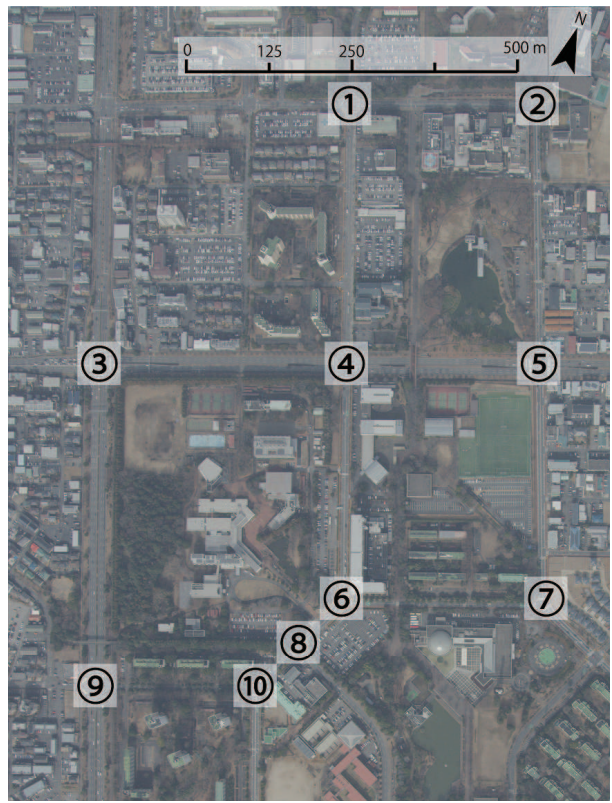}
  \caption{Target area in the movie. Car crossing timing into the numbered intersections. Other traffic flows are not considered in this study.}
  \label{fig:map}
\end{figure}

The traffic prediction model proposed above can be applied to real-time traffic data as follows. 
The traffic data used in this study were collected from a movie shot from the sky by a hovering helicopter, which implies that the data is obtained from a bird’s eye view. The details on the movie shooting and data collection are summarised in Table \ref{tab:table}. 
A part of the area shot is shown in Fig. \ref{fig:map}, which is the centre area of Tsukuba city, Japan, and is considered as the target area for prediction. 
It is possible to obtain the timeline and location of all vehicles in the movie. 
By using the vehicle information, we can consider several prediction tasks of traffic dynamics from a partial observation of those moving vehicles. 
This partial observation and dynamics prediction should be reasonable as 
it is not realistic to obtain all vehicle information in any given area at any time in terms of observation cost. 
Therefore, we perform the prediction task with respect to a subset of the available traffic data. 
Specifically, we focus on the traffic flow at the intersections numbered in Fig. \ref{fig:map}. 
In this study, the traffic flow is defined as a series of timings at which a vehicle passes each stop line of the intersections from the in-coming direction. 
Moreover, we experiment with predictions about causally irrelevant parts of the traffic data, 
namely, predicting the input flow to the network from the inside flow, 
rather than traffic flows that can be predicted by simple temporal causality. 
In this study, the time series of the traffic volume 
at the entrance to the No. 5 intersection from the east is inspected. 
Clearly, it is not causally related to the traffic flow at any other numbered intersections in Fig. \ref{fig:map}.

It should be noted that the task above is not a specific case but an example. It would be possible to consider other tasks from the movie. We also remark that, in general, it is not cost-effective to keep shooting vehicle motion from the sky by helicopters. 
Therefore, it is important to know how much of the whole picture of vehicle motion in the area of interest can be captured by limited observations with low computational costs. 
There should be a trade-off between observation and computation; lower (higher) observation costs require higher (lower) computational costs to capture the whole picture. 

\begin{table}
 \caption{Details of the movie shot from helicopters and collected data.}
  \centering
  \begin{tabular}{lll}
    \toprule
    Date    &  10:49:38 $\sim$11:51:46, Feb. 14, 2020  (Weather: sunny) & \\
    &7:37:35 $\sim$ 8:32:37, Feb. 18, 2020  (Weather: sunny)& \\
    \midrule
    Movie & Shooting by an 8K camera at a 2,000 m altitude &      \\
    Shooting range  & 2 km north--south and 1 km east--west & \\
        Counted objects    &   All motor vehicles, including heavy vehicles and motorcycles, except for bicycles    &   \\
        How to count    & Recording the time when a vehicle crosses the stop line at the intersections &      \\
        Number of counts & 3600 in 60 mins on Feb. 14 and 3060 in 51 min on Feb. 18& \\
        Remark & Angle of view changes significantly when the helicopter turns \\
    \bottomrule
  \end{tabular}
  \label{tab:table}
\end{table}

\subsection{Methods of prediction}
First, we name the time series of the traffic volume at the intersections as `intersection Number'+`in-coming direction (from north (n)/ south (s)/ west (w) /east (e))', e.g. `1e', `5s', etc. 
By using this notation, our target for the prediction is `5e'.  
The time series of the traffic volume is obtained by summarising the passing timings of vehicles with a time interval; the number of timings within a fixed length of the time interval is the value of the time series at a time step. 
For simplicity, we assume that the value corresponds to the traffic volume $X_i$ in a link of the transportation network.  
We call this time interval as `interval', which determines the time scale of the time series. 
Another parameter for the prediction task is `$\tau$', which is the forecasting horizon. 
Furthermore, to improve the prediction performance of the proposed method, we introduce a multiplexing method \cite{Nakajima2018}, in which 
$X_i$ enlarges its dimension by adding a shifted time series. More specifically, 
$X_i=[x_i(t_0),x_i(t_0+1),\ldots ,x_i(t_0+T) ]^\top$ changes its shape as
\begin{align}
X_i=[[x_i(t_0),\ldots ,x_i(t_0+T) ]^\top \ [x_i(t_0+1),\ldots ,x_i(t_0+T+1) ]^\top \ \ldots [x_i(t_0+p),\ldots ,x_i(t_0+T+p) ]^\top],
\end{align}
 where $p$ is the amount of maximum shift. Note that $x_i$ is the observed data.
  The multiplexing method increases the dimension of $X$, which improves the prediction. 
Once the three parameters and the target time series are fixed, 
it is possible to calculate the output weight matrix $W_{out}$ by Eq. (\ref{traffic_comp}). 
The length of the training time series is assigned to the ratio $r$ of the whole available time series, and the rest is for the test.

To predict the target time series by a conventional method, 
the dynamics of the traffic flow must be modelled and simulated by computers. 
As for our scheme, the modelling and simulation correspond to observing the real-time series. 
This is a crucial point of CH, that is, 
real-world phenomena are simulating a model instead of digital computers. 
All the computer has to do is to estimate $W_{out}$. 
One remark is that the state of $\bm{X}$ does not necessarily include all the links for $X_i$. 
This provides resilience of CH.

For the quality evaluation of the prediction, we adopt the normalised root mean square error (NRMSE) between the actual data $y_t$ and the predicted data $\hat{y}_t$, i.e. $\sqrt{\frac{\sum(\hat{y}_t-y_t)}{\bar{y}}}$, where $\bar{y}$ is the standard deviation of $y_t$. 
In addition to the prediction error, 
we consider the distance between the attractors in a delay coordinate space reconstructed from the time series $\{y_t\}$ and $\{\hat{y}_t\}$ \cite{Takens1981}. 
This comparison of distances is intended to measure how much the model $W_{out}$ learns the dynamical structure: $\dot{y}=F(\bm{X}, y) $, where $F$ generates the target time series. 
In this study, we assume that the reconstructed attractors are on a lower dimensional manifold and that the distance between the manifolds related to the target time series and the predicted one is calculated in the delay coordinate space. We adopt the Wasserstein distance (WD) between two attractors \cite{Muskulus2011}. For simplicity, the reconstructing dimension is $3$, and the delay is one time step; i.e., the coordinate of the space is $(t,\ t-1,\ t-2)$.

The baseline method is the prediction with the autoregressive (AR) model using the data of `5e' itself. Here, the AR model is formulated as follows:
\begin{align}
y_t=c_t+\sum_{k=t-p}^{t-1}c_ky_k, 
\end{align}
 where $c_*$ are the parameters estimated from training data.
Note that the data of `5n, 5s, and 5w' are not used in the proposed prediction 
as the periodicity of the traffic signal at intersection 5 significantly affects the dynamics of `5n, 5s, 5w, and 5e'. Therefore, the dynamics of `5e' strongly depends on `5n, 5s, and 5w'.

In addition to the one-shot estimation of $W_{out}$ for the training time series, 
we consider a real-time update of $W_{out}$; i.e., the weight matrix calculated at time $t$ is denoted by $W_{out}(t)$, which is updated by available time series. 
As the length of the training time series increases with time, 
the precision of the real-time prediction should improve. 
To compare one-shot $W_{out}$ with real-time $W_{out}(t)$, 
we define two parameters $r_{1}$ and $r_{2}(\ge r_1)$. The former is the ratio of the length of the training time series to the whole time series $T$. The latter is the ratio of the length of the test time series to the training one, which means that $\Delta=(\frac{1}{r_2}-1)r_1T$ is the length of the test time series for the updated $W_{out}(t)$. First, $W_{out}(1)$ is estimated by the first $r_1T$ time series. 
By $W_{out}(1)$, $\Delta$ length time series following the training one is predicted. 
Thereafter, in the next step, $T_1=r_1T+\Delta$ is the length of the training time series for estimating $W_{out}(2)$, and, again, $\Delta$ is the test length for the updated $W_{out}(2)$. We repeat the update of $W_{out}(t)$ until the total length of the prediction time series is greater than $(1-r_1)T$. Note that the one-shot estimation of $W_{out}$ corresponds to $r_{1}=r_2$.

\subsection{Results of prediction}

\begin{figure}
\begin{tabular}[c]{cc}
\begin{minipage}{.5\textwidth}
\begin{center}
  \includegraphics[width=1.2\textwidth]{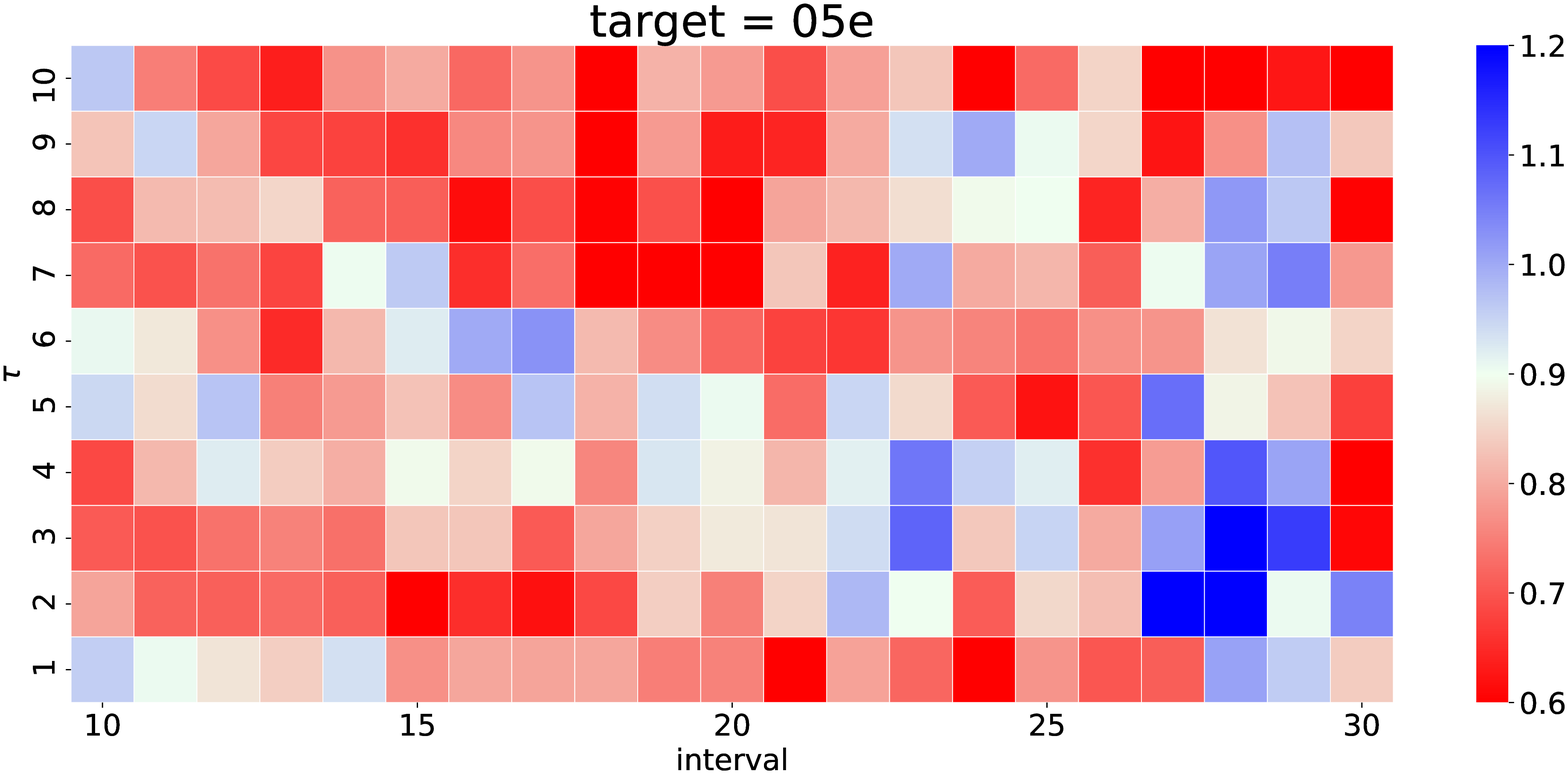}
  \hspace{1.6cm} {(a) Precision of the prediction on Day 1}
  \end{center}
  \end{minipage}   &
  \begin{minipage}{.5\textwidth}
\begin{center} 
  \includegraphics[width=1.2\textwidth]{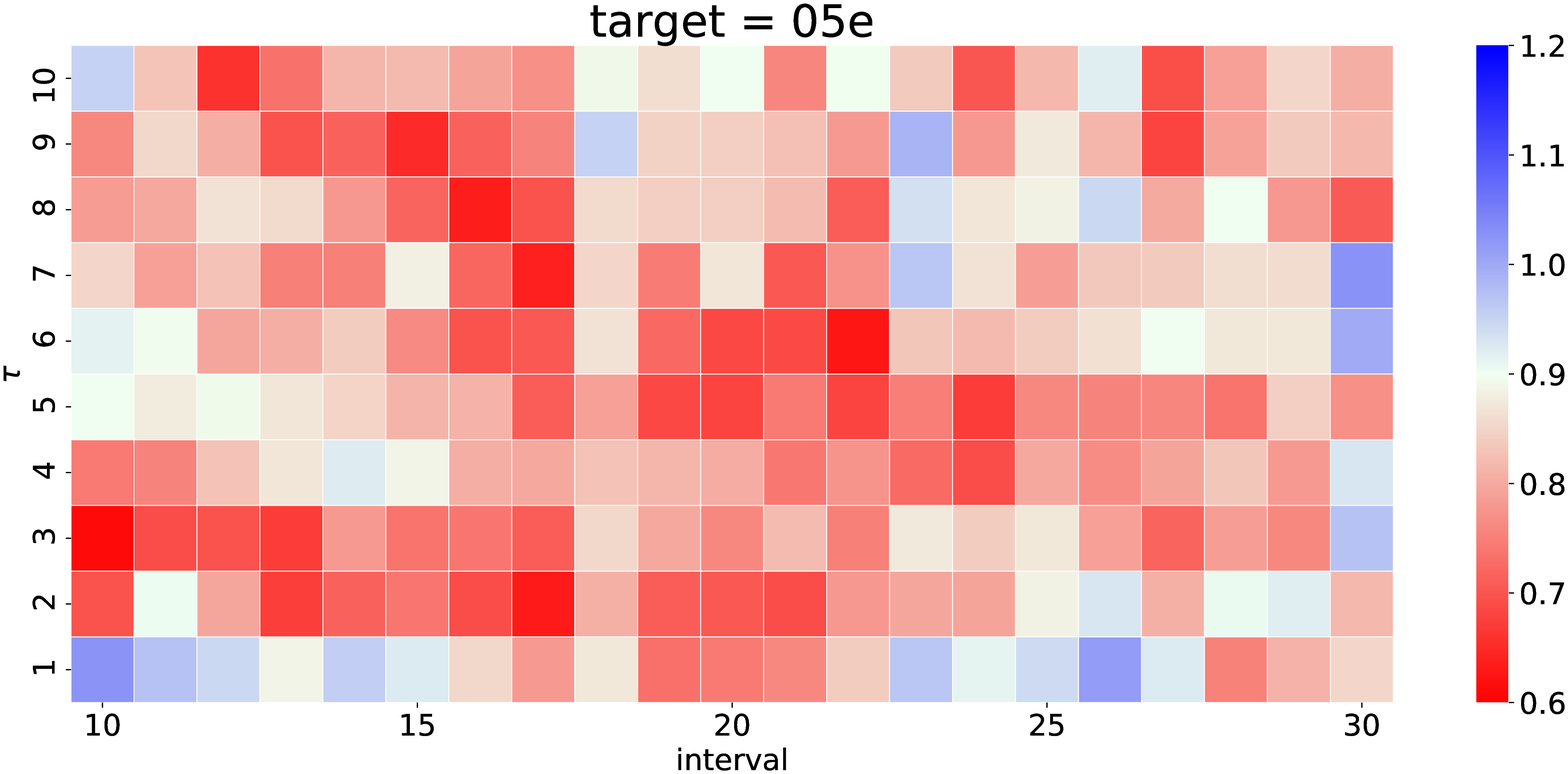}  
  \hspace{1.6cm} {(c) Precision of the prediction on Day 2 from the data of Day 1}
  \end{center}
  \end{minipage}  \\
    \begin{minipage}{.5\textwidth}
\begin{center} 
  \includegraphics[width=\textwidth]{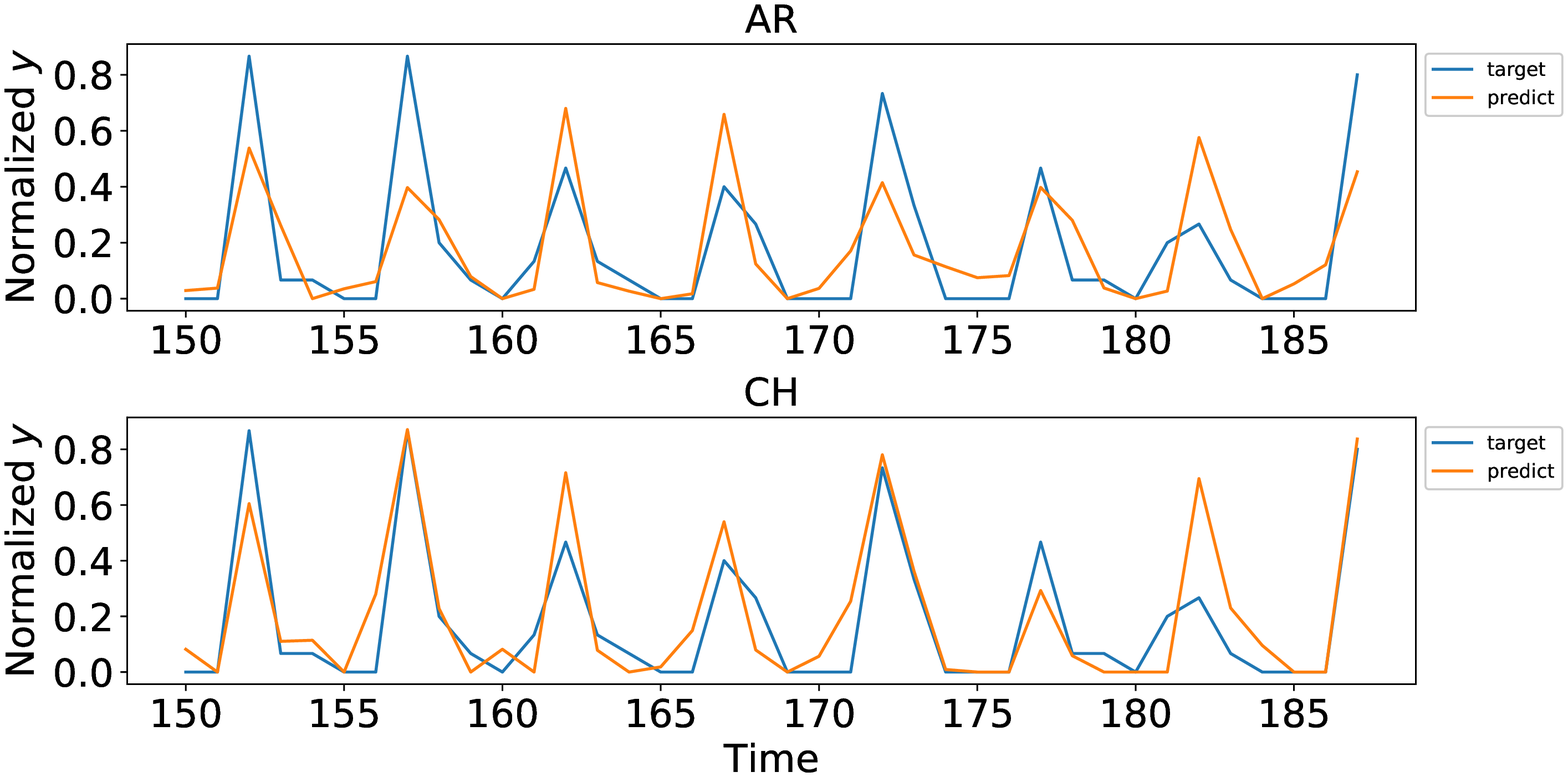}  
  \hspace{1.6cm} {(b) Predicted time series on Day 1. }
    \end{center}
  \end{minipage} &
    \begin{minipage}{.5\textwidth}
\begin{center} 
  \includegraphics[width=\textwidth]{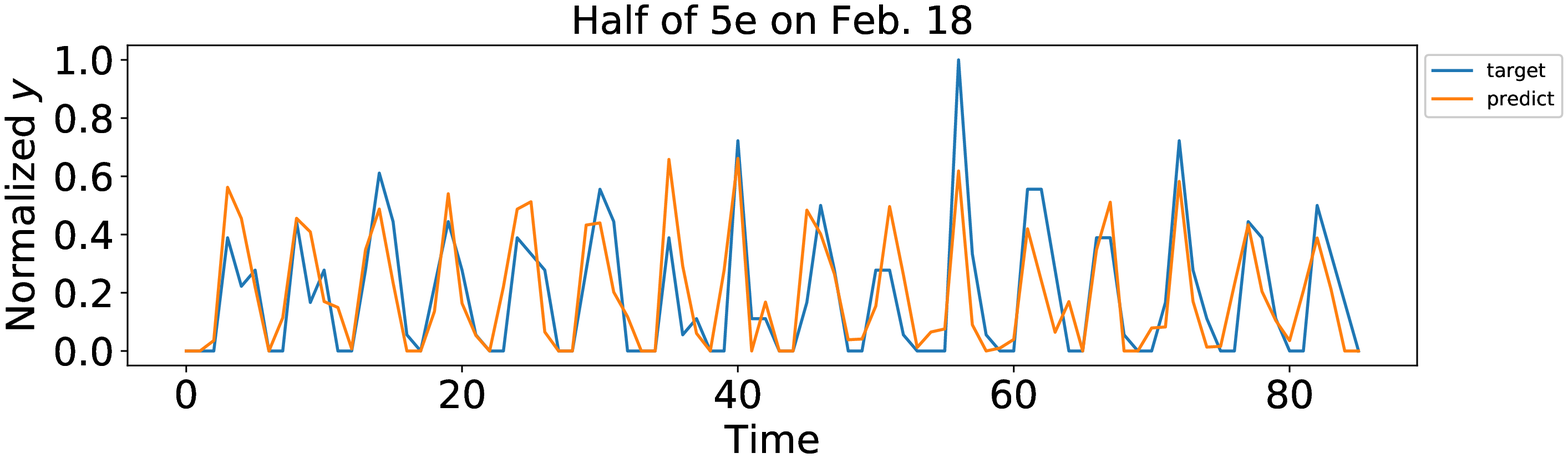}  
 \hspace{1.6cm} {(d) Predicted time series on Day 2 from the data of Day 1}
  \end{center}
  \end{minipage}
  \end{tabular}
  \caption{Precision of the time series prediction ((a), (c)) with respect to $\tau$ and interval and predicted time series ((b), (d)). The parameters are as follows: in (a) and (c), $p=2$; in (b), $\tau=7$, interval$=18, and p=6$; and in (d), $\tau=7$, interval$=17,$ and $p=2$. 
  The NRMSEs for AR and CH in (b) are 0.651 and 0.526, respectively.
  The time series shown in (d) is the half of the full length on Day 2. The time series is normalised to $[0,1]$. } \label{fig:pred}
\end{figure}

Figure \ref{fig:pred} shows the prediction error and the predicted time series for Feb. 14 (Day 1) and 18 (Day 2) when $W_{out}$ is learned from a part of the time series on Day 1. Figure (a) shows the prediction errors with respect to $\tau$ and the interval for $p=2, r=0.8$. The colour represents the NRMSE. 
Although the prediction errors increase in a limited region, the errors are within 0.6 for many parameters. This error value is acceptable for the prediction. 
In fact, the predicted time series are shown in (b) for the parameters $\tau=7, \text{interval}=18, p=6,$ and $r=0.8$.  
The upper panel shows the results obtained by the AR model with $p=6$, and the bottom panel shows that by the proposed method, i.e. the result of CH. The normalised RMSEs are 0.651 and 0.526, respectively. 
Figure (c) shows the prediction errors when the time series on Day 2 by $W_{out}$ estimated from that on Day 1 with $p=2$. 
The errors are within 0.7 for a large part of the parameters. The corresponding time series are shown in (d). 
This prediction on Day 2 is performed to validate the generalisation performance of CH. 
The results indicate that the expressiveness of the traffic dynamics is not necessarily unique only within one day but has some generalisation performance for another day. One of the reasons would be the periodicity of the time series, which depends on the period of traffic signals. However,  the amplitude of the oscillation can be also predicted somehow, which is not possible only from the periodicity.

Note that the AR model can perform similar to or better than the CH method with $p=1$ for a large parameter region of $\tau, \text{interval}$ as it uses the past data for the target time series. 
However, the CH method with multiplexing, i.e. $p\ge 2$, improves its prediction performance to the precision level of AR. 
Additionally, note that predicting the time series of Day 1 by $W_{out}$ of Day 2 is not more accurate than the results predicted from the data on Day 1. This observation indicates that traffic patterns on Day 1 have stronger generalisation than those on Day 2.

\begin{figure}
\begin{tabular}{ccc}
\begin{minipage}{.33\textwidth}
\begin{center}
  \includegraphics[width=\textwidth]{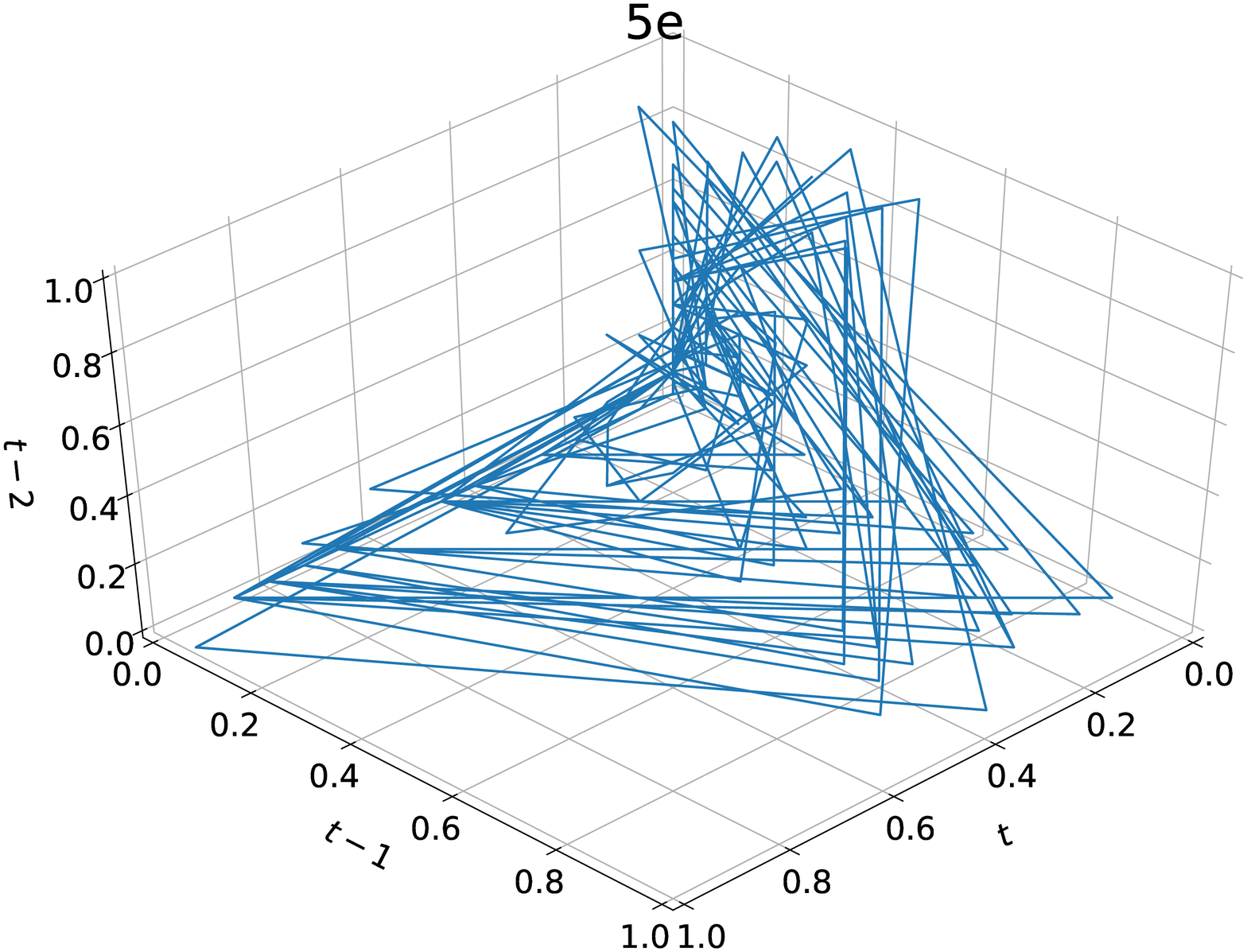}
  \hspace{1.6cm} {(a) Target: 5e.}
  \end{center}
 \end{minipage}   &
  \begin{minipage}{.33\textwidth}
\begin{center} 
  \includegraphics[width=\textwidth]{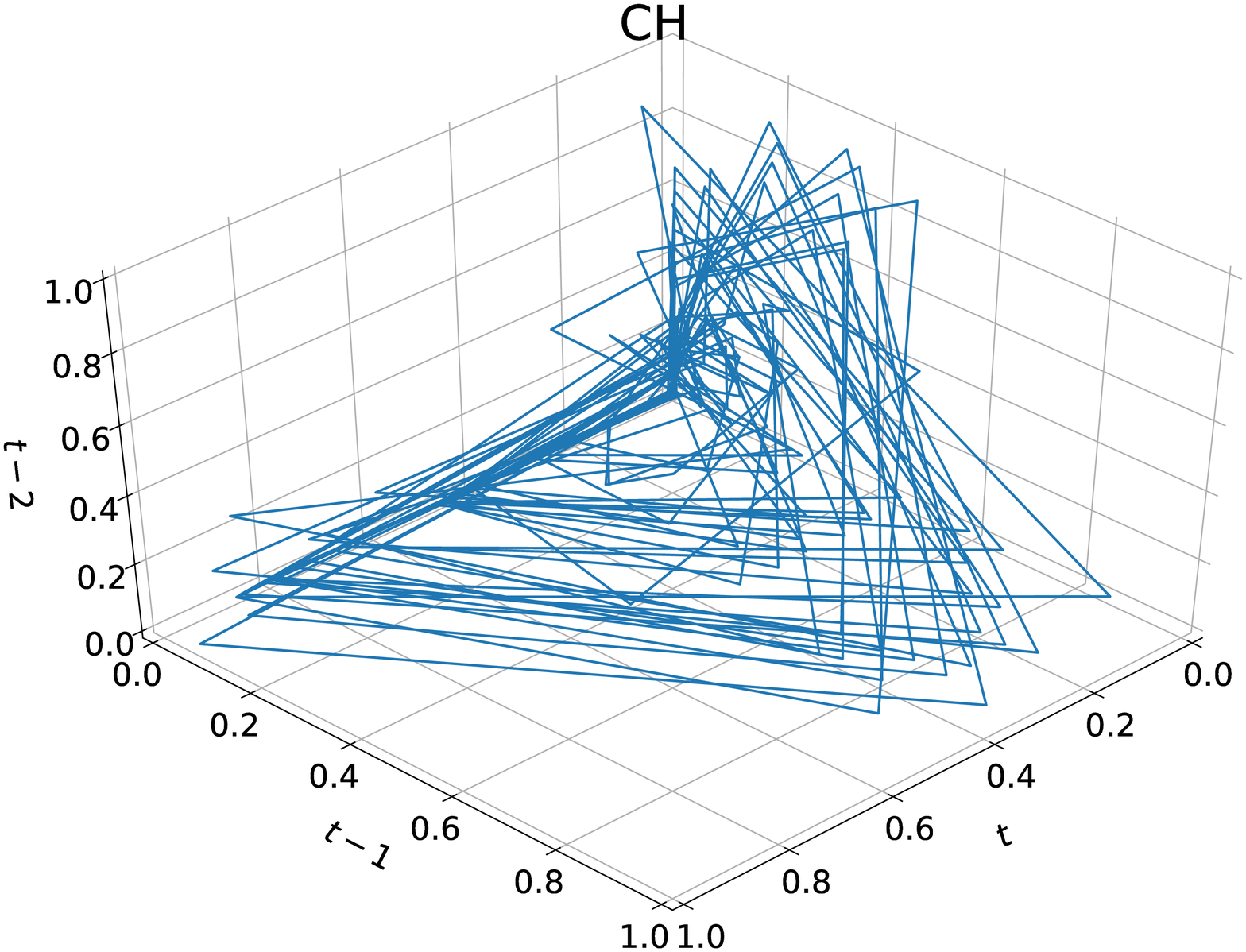}  
    \hspace{1.6cm} {(b) Predicted by the proposed model.}
  \end{center}
  \end{minipage}  &
    \begin{minipage}{.33\textwidth}
\begin{center} 
  \includegraphics[width=\textwidth]{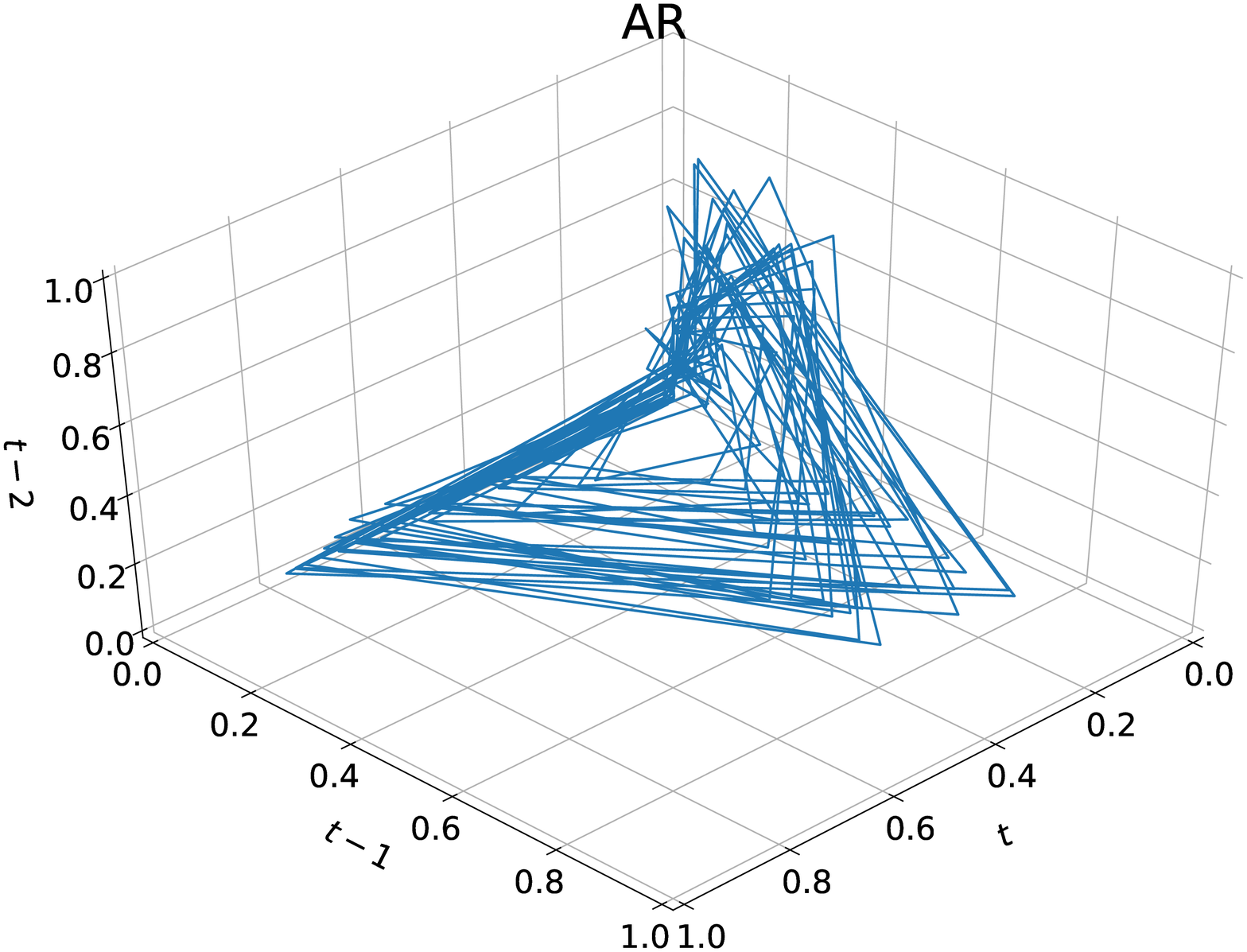}  
      \hspace{1.6cm} {(c) Predicted by the AR model.}
  \end{center}
  \end{minipage} 
    \end{tabular}
    \begin{center}
    \begin{minipage}{.5\textwidth}
\begin{center} 
  \includegraphics[width=\textwidth]{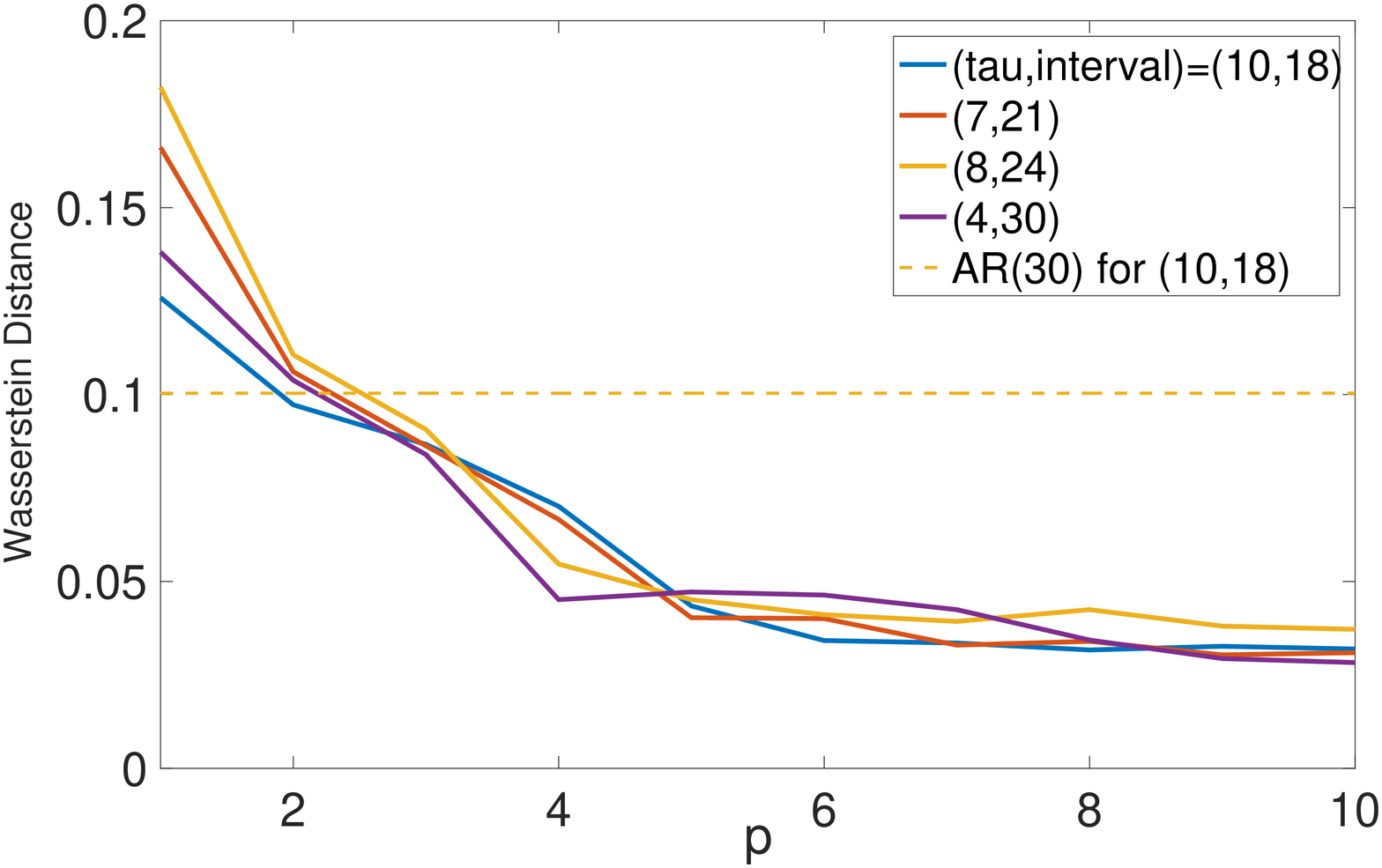}  
        \hspace{1.6cm} {(d) Wasserstein distance with respect to $p$.}
  \end{center}
  \end{minipage}
      \end{center}
  \caption{(a,b,c) Reconstructed attractors for the test time series on Day 1 with $\tau=7$, interval$=18,$ and $p=6$. (d) Wasserstein distance between the target attractor and the predicted one depending on $p$ for several pairs of $(\tau, \ \text{interval} )$. The dashed line shows the distance for the attractor of the AR model with $p=30$.} \label{fig:attractor}
\end{figure}

Next, we measure the distance of the attractors reconstructed in delay coordinates. Figures \ref{fig:attractor} (a), (b), and (c) show the reconstructed attractors for the target, predicted by the CH and AR models. $p=6, \tau=10,$ and $\text{interval}=18$ for CH, and $p=6$ for the AR model. 
The structures of the reconstructed attractors look similar. 
The normalised RMSEs for (b) and (c) are 0.676 and 0.558, respectively. 
The WD to the target from CH is 0.0342, and the WD from the AR model is 0.0850. 
Furthermore, we investigate the WD with respect to $p$, as shown in Fig. \ref{fig:attractor} (d). 
Results for several different pairs of parameters $(\tau, \text{interval})$ for CH are shown. 
For comparison, we consider $p=30$ for the AR model. The WD from CH decreases with increasing $p$ for each parameter pair. In addition, it becomes lower than the WD from the AR model. 
The results may indicate that the higher dimensional time series except for `5e, 5w, 5n, 5s' are sufficiently complex, which makes it possible to capture the structure of nonlinear dynamics behind `5e' by a linear model. It should also be remarked that the periodicity of each time series is effective for prediction performance. In other words, the prediction could be considered as a linear combination of the basis functions represented by the time series, whose periods are correlated with the target one.

\begin{figure}
  \centering
  \includegraphics[width=\textwidth]{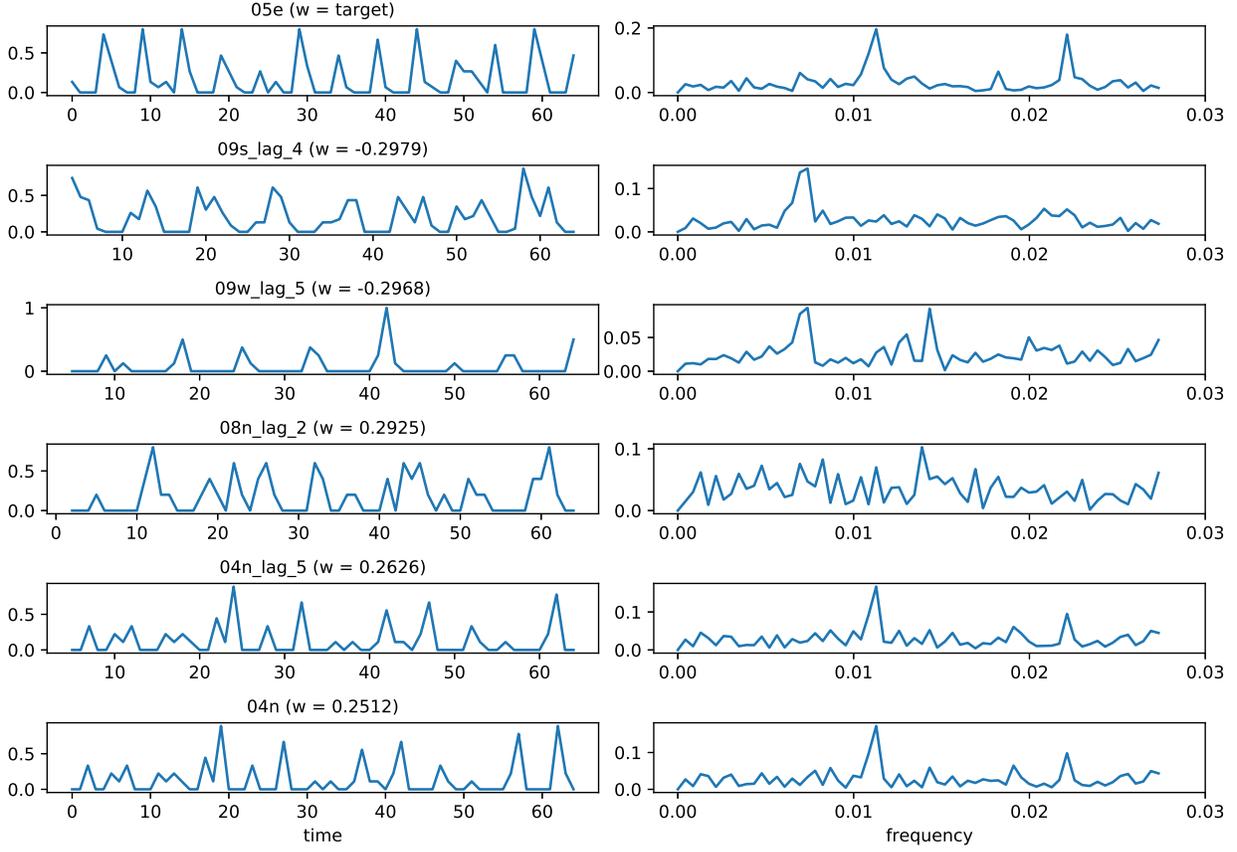}
  \caption{Time series that compose the prediction in Fig. \ref{fig:pred} (b) and its power spectrum. The top panel shows the target time series, and the others are sorted in the order of the absolute values of $W_{out}$ toward the bottom. $x$ in lag\_$x$ is the amount of time shift. $\tau=7$, interval$=18,$ and $p=6$.}
  \label{fig:FFT}
\end{figure}

\begin{figure}
\begin{tabular}{ccc}
 \begin{minipage}{.5\textwidth}
\begin{center}
    \includegraphics[width=\textwidth]{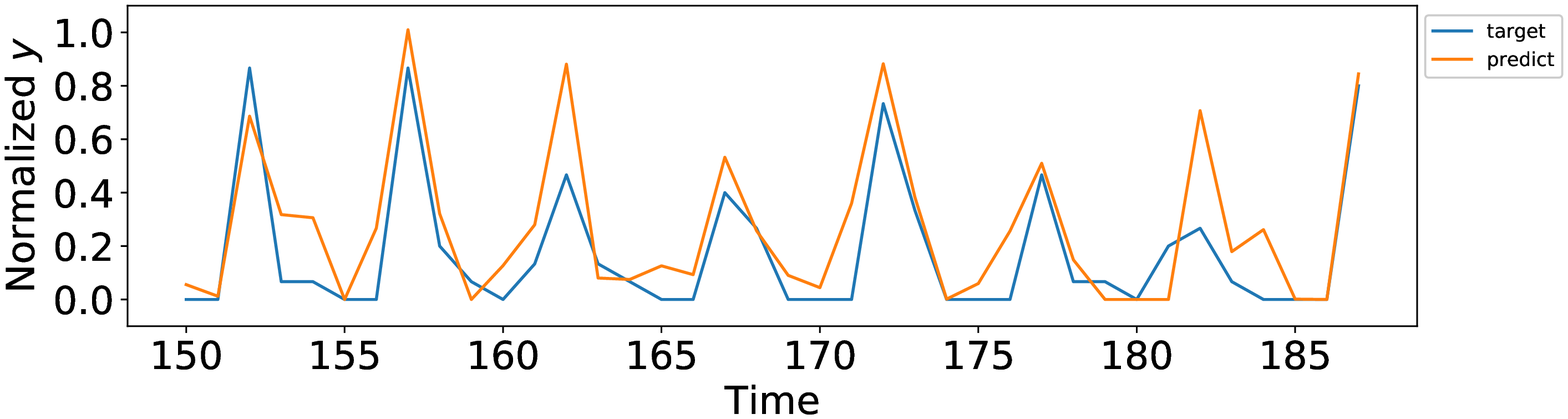}
    \hspace{1.6cm} {(a) Fixed $W_{out}$. NRMSE: 0.6579}
  \end{center}
 \end{minipage}   &
 \begin{minipage}{.5\textwidth}
\begin{center}
  \includegraphics[width=\textwidth]{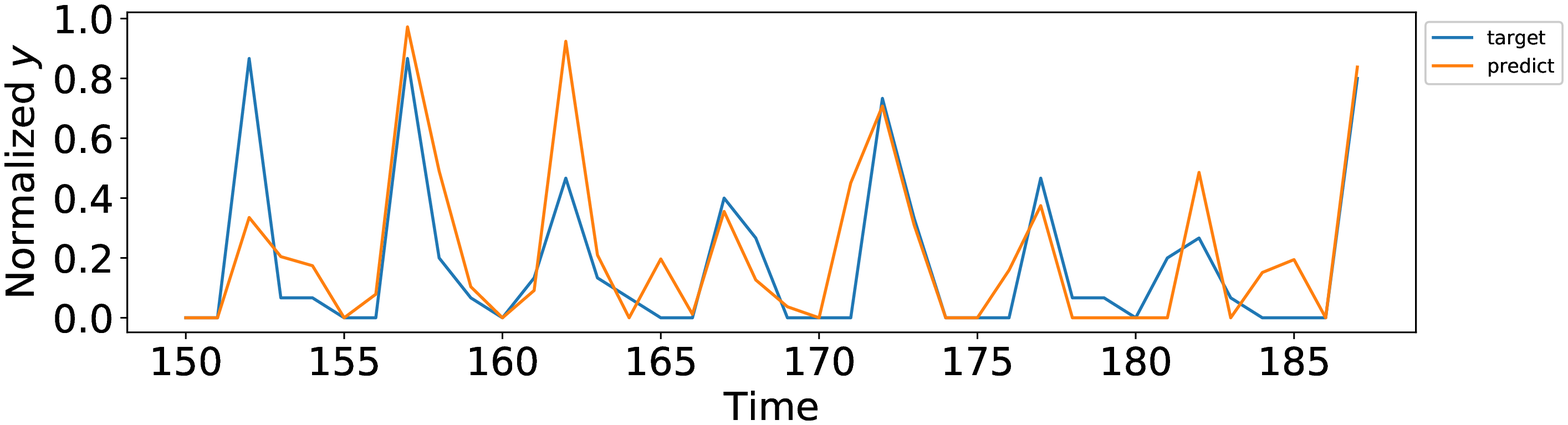}
  \hspace{1.6cm} {(b) Re-learned $W_{out}$. NRMSE: 0.6598}
  \end{center}
 \end{minipage}   
 \end{tabular}
  \caption{Time series predicted by a part of the observed time series in Fig. \ref{fig:map}. The panels   
 show the predicted time series after removing time series '9w', '9s', and '8n' considering the power spectrum in Fig. \ref{fig:FFT}. (a) $W_{out}$ is learned before removing the time series but with the corresponding parts deleted. (b) $W_{out}$ is learned after removing the time series. $\tau=7$, interval$=18,$ and $p=6$.}
  \label{fig:robust}
\end{figure}

Here, we show what type of computation is conducted by harvesting in the present example. 
Figure \ref{fig:FFT} shows a part of linearly combined time series (left column) as well as their 
corresponding power spectra in the right column. The top panels show the target, and the other panels are ordered by the amplitude of the absolute value of $W_{out}$ for the top five.  
As can be seen from the power spectra, the target time series are composed of the time series whose distribution of the power spectra are similar to that of the target. 
Therefore, in the harvested computation, it could be said that Fourier series expansion of the target time series is conducted by spatially distributed traffic dynamics together with taking fluctuations in the amplitudes of the time series into account. 
Furthermore, to ensure the resilience of the harvesting combination of the time series, we remove some of the significant time series in terms of power spectrum in Fig. \ref{fig:FFT}. There are two ways of removing the time series for evaluating resilience. The first option is that $W_{out}$ is learned from all time series, and we remove some of the components from $\bm{X}$ and $W_{out}$. The second option is to learn $W_{out}$ after removing the time series. The predicted time series from the reduced time series are shown in Fig. \ref{fig:robust}. 
Note that the predictions are resilient against the patterns of the removed time series (not shown). Those results indicate that there are several combinations of the time series for composing the target time series, which implies that the harvestable computation is embedded in parallel.

\begin{figure}
\begin{tabular}{ccc}
\begin{minipage}{.33\textwidth}
\begin{center}
  \includegraphics[width=\textwidth]{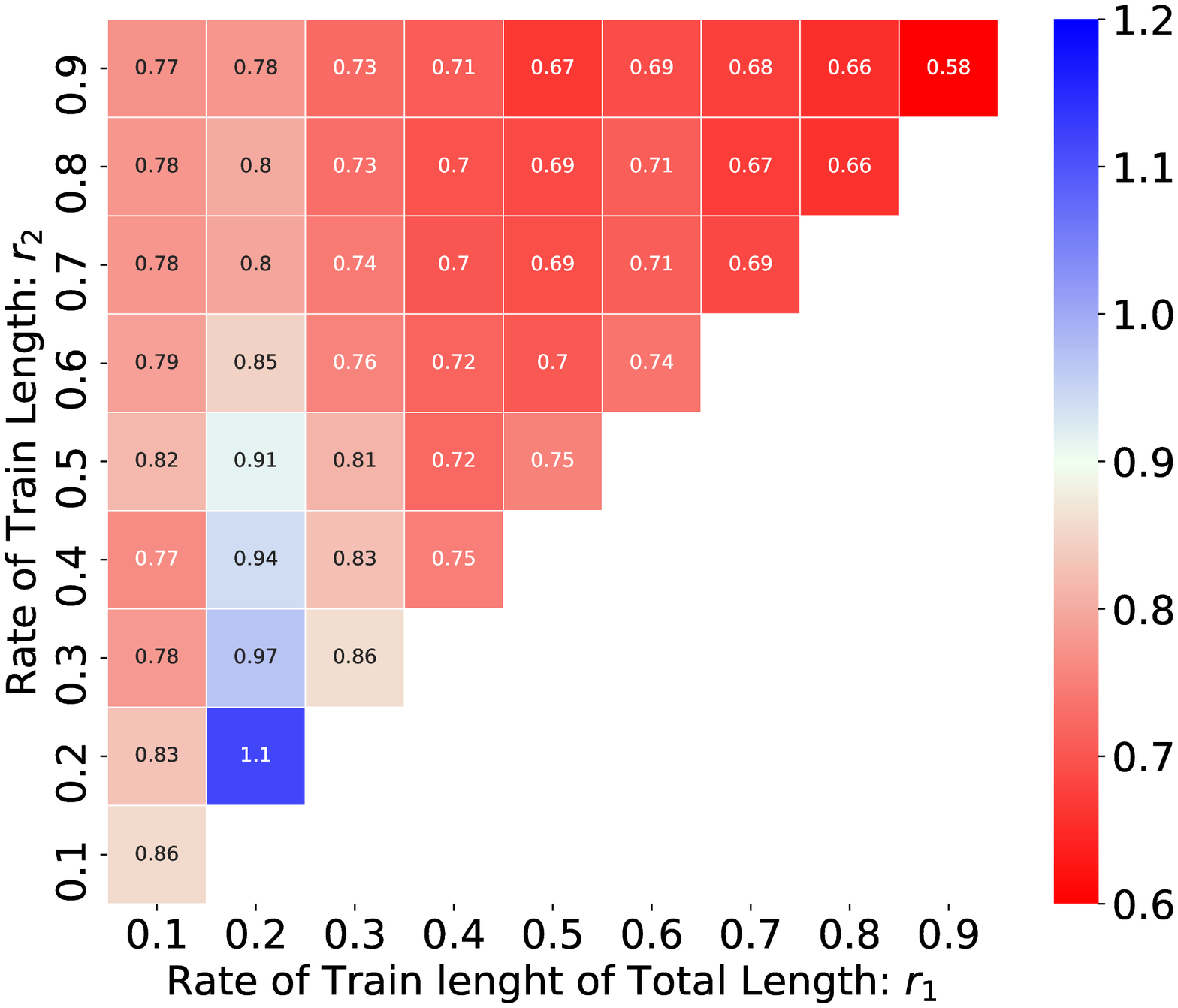}
   \hspace{1.6cm} {(a) Real-time prediction for Day 1.}
  \end{center}
  \end{minipage}   &
  \begin{minipage}{.33\textwidth}
\begin{center} 
  \includegraphics[width=\textwidth]{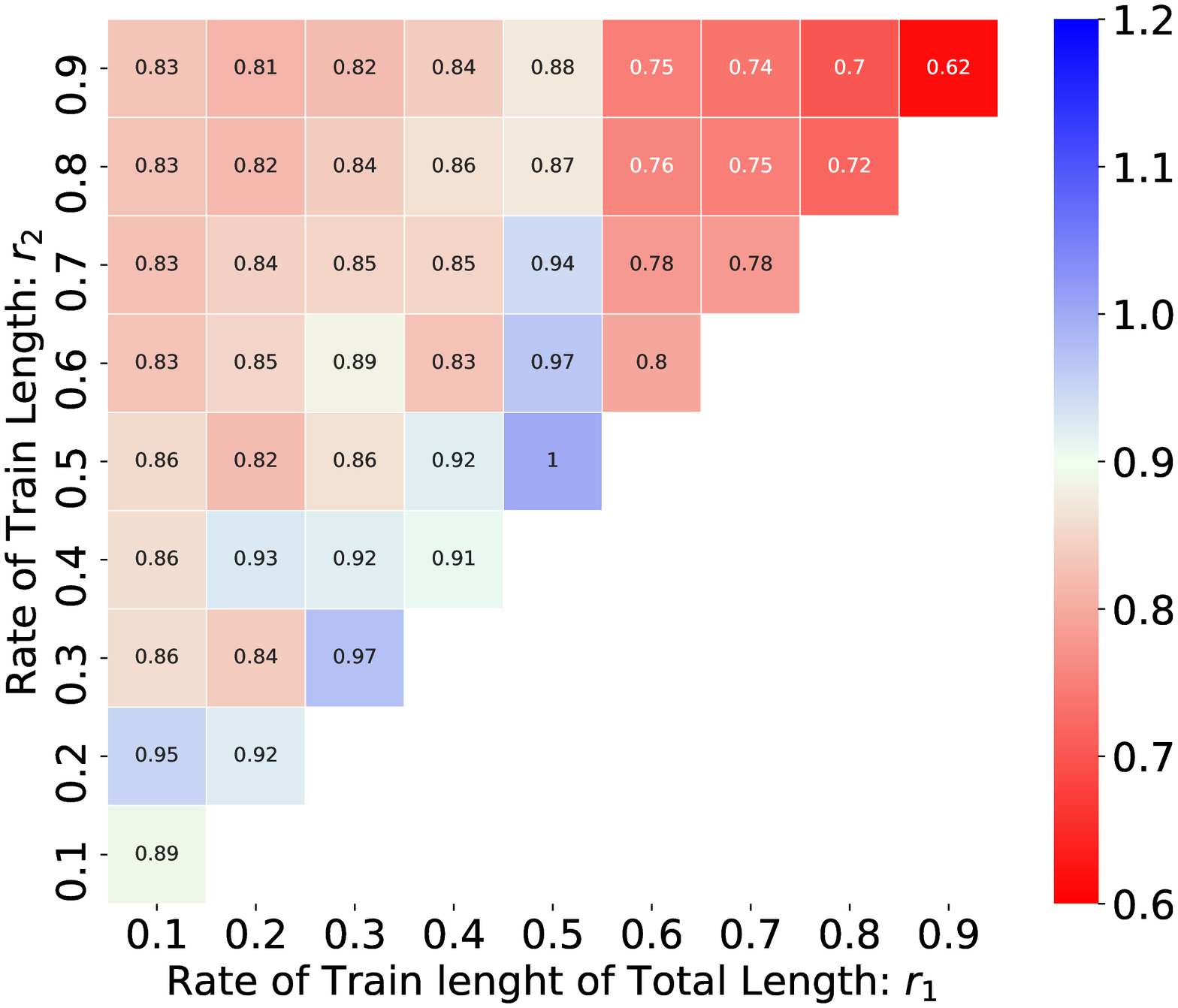}  
  \hspace{1.6cm} {(b) Real-time prediction for Day 2.}
  \end{center}
  \end{minipage}  &
    \begin{minipage}{.33\textwidth}
\begin{center} 
  \includegraphics[width=\textwidth]{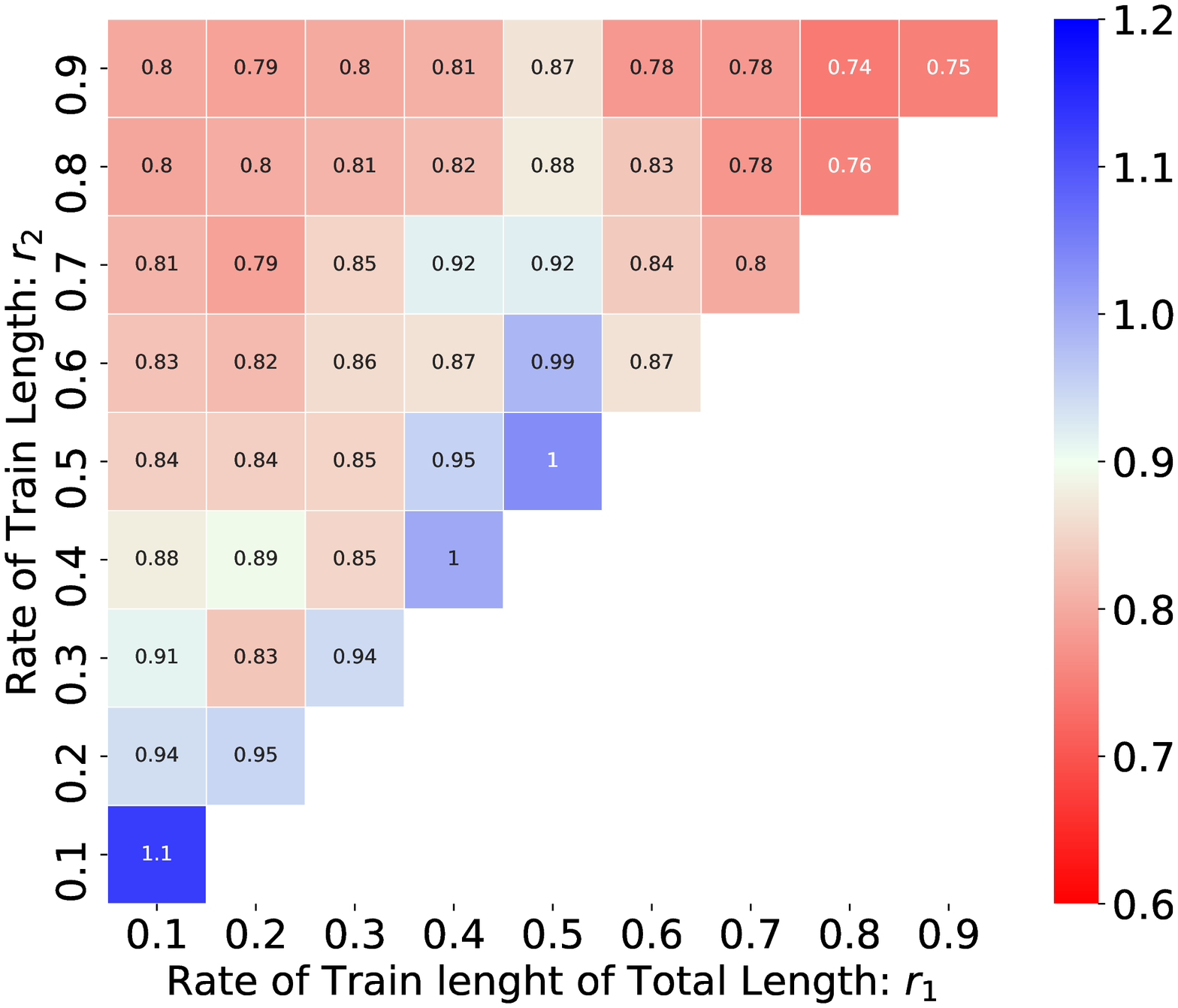}  
  \hspace{1.6cm} {(c) Real-time prediction for Day 2 added to Day 1.}
  \end{center}
  \end{minipage}  
  \end{tabular}
  \caption{Precision of the time series prediction with the real-time update of $W_{out}(t)$. $\tau=7, \text{interval}=18$.  
  (a) Prediction within the time series of Day 1. 
  (b) Prediction of the time series of Day 2 combined after Day 1.  } \label{fig:online}
\end{figure}

Finally, Fig. \ref{fig:online} shows the prediction errors for the target time series predicted by updating $W_{out}(t)$. 
In figures (a) and (b), the target time series are only of Days 1 and 2, respectively. 
Figure (c) includes the time series of Day 2; the target time series is the time series that combines Day 2's after Day 1's. 
As explained before, $r_1=r_2$ corresponds to the one-shot estimation of $W_{out}$, which is the diagonal line in the figure. 
Thereafter, below the diagonal line with fixed $r_1$, the prediction errors become smaller towards large $r_2$. 
Intuitively, when $r_1$ is fixed and $r_2$ is large, small $\Delta$ or a frequent update of $W_{out}(t)$ makes the prediction better. 
This tendency holds for all the cases from (a) to (c). 
Furthermore, from the comparison of figures (b) and (c), on one hand, large $r_1$ improves the prediction within Day 2, as compared to the addition of Day 1. On the other hand, small $r_1$ with large $r_2$ provides better prediction on Day 2 added to Day 1 than only on Day 2. 
The former case implies that if one has sufficient information on the traffic dynamics of a single day, one can have sufficiently good prediction on that day. The latter case means that if the available information is not sufficient for a single day, the information on the other day provides better prediction than using only the information of the single day.

\section{Summary and Discussion}

In this paper, we proposed a concept of CH with real-world data and demonstrated its feasibility with traffic dynamics. 
Specifically, we focused on a task of the prediction of traffic volume at the intersections, where the 
target time series can be predicted by a network of the time series at the intersections apart from the target. We performed sufficiently precise prediction in terms of normalised RMSE. 
Compared to the prediction by the AR model, the prediction errors were even. To evaluate the expressiveness of nonlinear dynamics represented by the network time series, we reconstructed the attractors in delayed coordinates and measured the distance to the target one. As a result, the proposed method learned the attractor closer to the target, as compared to the AR model. In addition, owing to the efficiency in computational cost of the proposed method, a methodology for real-time updating the prediction parameters is also proposed, and its effectiveness is verified. 
We discuss the characteristics of CH as follows.

First, the CH approach is well suited for the current state of the IoT technology, which provides unwieldy data collected from many sensors, because of its lower implementation cost. There exist related works with a similar motivation, such as in-network computation in wireless sensor networks by mapping artificial neural networks \cite{Pascale2018} and reservoir computing with impulse-based wireless sensor networks \cite{Peper2016,Kamei2018}. Our focus of CH can include these computing frameworks when considering wireless sensor networks as an existing infrastructure. 
In other words, our main claim in CH is that we reduce computational cost and designing cost for resources as much as possible. It should be noted that although the movie data used in this study were shot from the helicopter, the used data can be collected by a traffic counter at the intersections. These data could be observed by existing infrastructures. Furthermore, we emphasize that the proposed framework is free of implementation costs, designing costs, and sensor failures as many combinations of existing sensors can perform the target computation, as shown in Fig. \ref{fig:robust}, namely, it is a resilient computing problem.  

Furthermore, we discuss the rationale behind the superior computational and predictive performance exhibited by CH. There exists an efficient artificial neural network architecture that can solve a task without being trained as efficiently as a heavily trained deep neural network (see, weight agnostic neural networks (WANN) \cite{GaierHa}). Like the WANN, CH may utilise a network inherent to a dynamical phenomenon with a desirable architecture that already exhibits good predictive performance. CH seems resilient against disturbances in maintaining the predictive performance. In fact, removing the dominant top-ranked spectra still retains the predictive performance (Fig. \ref{fig:robust}). This observation indicates that the good predictive performance of CH might be supported by various spectra that are often redundant rather than a few top-ranked features. Redundancy confers a buffering effect against sudden component loss via a failure. In general, redundancy is one of the essential features in characterising the resilience of a given system against disturbances \cite{kitano2002}. The existence of redundancy in the power spectra for CH can lead to resilient prediction.

Next, we compare CH with alternative methods for time series prediction. 
As already mentioned in the results, the AR model is one of the simplest methods for the comparison with the proposed method. With regards to the data used in this study, the AR model has good prediction performance, and the proposed method outperforms the AR model in some cases. This may be because the time series network can learn the structure of a non-linear system generating the target time series, and some parts of the structure cannot be learned by the target time series alone. 
Moreover, our model could potentially outperform the AR model in terms of time series prediction after further improvements \cite{autores}. 

Another possible time series prediction method using big data is deep learning. The proposed method only learns a limited number of parameters within a limited time, so that the computational cost for the parameter estimation is considerably lower than that of deep learning. This implies that the proposed method is more suitable for real-time prediction. In fact, the effectiveness of real-time updating is verified with real-world data, as described above. Therefore, it is expected that the proposed method can replace deep learning 
when fast computing is required and the prediction precision need not be high. Furthermore, data assimilation is one of the prediction methods that use real data and performs prediction with high precision. However, it requires a detailed model for the target. As for the proposed method, no model and noise seeds are present as it uses real-world phenomena as they are. 

Third, in terms of understanding the mathematical properties of CH, machine learning with nonlinear dynamical systems such as reservoir computing would be helpful. For example, 
auto reservoir computing, in which internal time series of the target system are regarded as reservoir units, has recently been proposed \cite{autores}. This method uses the notion of spatiotemporal information (STI), which is mathematically equivalent to our scheme of parameter estimation. Therefore, analysing network of the time series with the prediction parameters in terms of STI can be a future work. Additionally, the auto reservoir exploits a neural network in addition to a linear combination of the time series. Consequently, the prediction performance is better than that of other conventional methods for several tasks. 
Our approach is based on the idea how to extract computational processes performed by real-world systems in the background. Accordingly, we focused on the prediction tasks based on linear combinations of real-world data. Therefore, a nonlinear transformation can improve the prediction performance of our method. 

In addition to predicting the internal state of the system, the computation is not necessarily limited to the harvested system. 
Therefore, it is possible to consider a generalisation of harvested computation by taking external inputs into account. For instance, with respect to predicting an external input, external time series prediction can be verified by a numerical simulation of road traffic dynamics \cite{ando2019}. 
Note that it is possible to consider the dynamics of road traffic in this study as a reservoir; i.e., it is possible to describe the dynamics by the echo state network (ESN)-based formulation. This is explained in the Appendix. 

Finally, in the proposed framework, only a linear combination of real-time series is considered. In this way, we can evaluate which time series are more effective for prediction. Using the example of traffic prediction, we explain how and why some components of the linear combination are effective for the prediction when a physical model of the traffic dynamics is established. 
Therefore, as a future work, we should consider establishing a physical model of real-world phenomena used for CH, which may be connected to explainability of computation tasks as well as computational universality of the target phenomena as a computing system.

\section*{Acknowledgement} 
This work was partially supported by JST MIRAI No. JPMJMI19B1 and JSPS KAKENHI No. 19K12198 and 17H03280. 
The data collection is supported by Tsukuba Smart City Council as well as MLIT Japan.
We are also grateful to the member of Urban Analytics Lab. in University of Tsukuba for their efforts in compiling the data.  

\section*{Appendix}

We assume a multi-agent model of traffic in a transportation network. 
For simplicity, the network is assumed to be in a directed lattice. 
In addition, each node of the network has traffic signals with periodic alternation. 
The periods of those traffic signals can provide complex dynamics to the traffic flow over the network. 
To introduce the principle of reservoir computing to the traffic model, 
we compare the formulation of the model and that of the ESN, which is the fundamental model of reservoir computing, as follows. 

Based on the description of the ESN \cite{Jaeger2004}, a traffic reservoir computer is formulated as follows: 
\begin{equation}
X(k+1)=f(WX(k)+W_{in}u(k)), \label{esn_traffic}
\end{equation}
where $X(k)$ is the number of vehicles at each link of the lattice network at discrete time $k\in \mathrm{N}$. 
The dimension of $X(k)$ is $2N$, which is the number of links in the directed lattice network with two links in both directions between two nodes. 
These two directions mean that there is at least one lane in each direction of the road. 
$W\in \mathrm{R}^{2N \times 2N}$ is a random transition matrix at each node from one in-coming direction to 
three out-going directions, and $W_{ij}$ is the transition probability from in-coming link $i$ to out-going link $j$, 
which is the $(i,j)$ component of $W$. 
Note that $W$ has at most three non-zero components in one raw and one column. 
$W_{in}$ corresponds to the weight matrix for external input $u(k)$ to the network, and, 
for simplicity, $W_{in}=0$ for the current system to preserve the number of vehicles.

$f$ is a nonlinear function, which is governed by traffic signals; 
that is, if $\pi>\mod(\theta,2\pi)>0$, then $f(x)=x,$; else, $f(x)=0$. 
In other words, the former corresponds to `{\it go}' signal, and the latter is `{\it stop}'. 
$\theta$ is an internal state of traffic signals, and its dynamics can be described as $\dot{\theta}=\tau$ doe periodicity. $f$ is applied to the element of the matrix. 
Note that $\tau$ is the time constant of traffic signals, and it can be varied with respect to signals. 
From this description (\ref{esn_traffic}), the traffic dynamics on a network is interpreted as a type of ESN, 
which implies that real-world traffic dynamics is considered as a reservoir. 

\bibliographystyle{unsrt}  
\bibliography{biblio}  

\end{document}